\documentclass[11pt,a4paper,hyphens]{article}
\usepackage[hyperref]{AACL-IJCNLP2020}
\usepackage{times}
\usepackage{latexsym}

\usepackage{microtype}

\usepackage[nolist]{acronym}
\usepackage{graphicx}
\usepackage{listings}
\usepackage{multirow}
\usepackage{booktabs}
\usepackage{pgfplots}
\usepackage{caption}
\usepackage{subcaption}
\usepackage{interval}
\intervalconfig {
soft open fences
}
\usepackage[htt]{hyphenat}
\usetikzlibrary{patterns}

\aclfinalcopy 
\def\aclpaperid{5} 

\newcommand{\en}{EN}
\newcommand{\thai}{TH}
\newcommand{\ms}{MS}
\newcommand{\id}{ID}
\newcommand{\hi}{HI}

\title{A Parallel Evaluation Data Set of Software Documentation\\with Document Structure Annotation}

\author{
Bianka Buschbeck \and Miriam Exel\\
SAP SE\\
Dietmar-Hopp-Allee 16, 69189 Walldorf, Germany\\
\texttt{\{bianka.buschbeck,miriam.exel\}@sap.com}
}

\date{}

\begin{document}
\maketitle

\begin{abstract}
This paper accompanies the \emph{software documentation data set for machine translation}, a parallel evaluation data set of data originating from the \emph{SAP Help Portal}, that we released to the machine translation community for research purposes.
It offers the possibility to tune and evaluate machine translation systems in the domain of corporate software documentation and contributes to the availability of a wider range of evaluation scenarios.
The data set comprises of the language pairs English to Hindi, Indonesian, Malay and Thai, and thus also increases the test coverage for the many low-resource language pairs.
Unlike most evaluation data sets that consist of plain parallel text, the segments in this data set come with additional metadata that describes structural information of the document context.
We provide insights into the origin and creation, the particularities and characteristics of the data set as well as machine translation results. 
\end{abstract}

\begin{acronym}
\acro{MT}[MT]{machine translation}
\acro{NMT}[NMT]{neural machine translation}
\acro{CAT}[CAT]{computer-assisted translation}
\end{acronym}

\section{Introduction}

The \emph{software documentation data set for machine translation} is created by SAP\footnote{\url{https://www.sap.com}} as evaluation data for the \ac{MT} research community. 
The data originates from the \emph{SAP Help Portal}\footnote{\url{https://help.sap.com}} that contains documentation for SAP products and user assistance for product-related questions. 
The current language scope is English (\en) to Hindi (\hi), Indonesian (\id), Malay (\ms) and Thai (\thai). 
The data has been processed in a way that makes it suitable as development and test data for machine translation purposes. 
For each language pair about 4k segments are available, split into development and test data. 

The segments are provided in their document context and are annotated with additional metadata about the document structure. The metadata provides information such as document and paragraph boundaries as well as the segment's text type, for example whether it is a title or table element. Such information will surely be valuable when developing and evaluating document-level \ac{MT} or for tuning systems for specific text types in order to increase the overall translation quality in this domain.

The \emph{software documentation data set for machine translation} as described in this paper is available under the Creative Commons license Attribution-Non Commercial 4.0 International (CC BY-NC 4.0). 
It is available on GitHub under \url{https://github.com/SAP/software-documentation-data-set-for-machine-translation}. 
It has been released by SAP for the 7th Workshop on Asian Translation (WAT 2020).\footnote{\url{https://lotus.kuee.kyoto-u.ac.jp/WAT/WAT2020/index.html}}

We will first provide some context, explaining the role of test data in machine translation and referring to related work (Section~\ref{sec:context}). We will then describe the origin of the \emph{software documentation data set} in Section~\ref{sec:origin}, including the data preparation and data selection. Section~\ref{sec:characteristics} is dedicated to the characteristics of the data set. Benchmarking results of \ac{MT} systems on the test sets are provided in Section~\ref{sec:mt}. Section~\ref{sec:concl} concludes.
\section{Context}
\label{sec:context}

Test sets are typically used for comparison in \ac{MT} evaluation campaigns, such as WMT\footnote{Yearly \emph{Conference on Machine Translation}, hosting a number of shared tasks. See \url{http://www.statmt.org/wmt19/} and the findings paper \citep{wmt19} for the 2019 occurrence.} and WAT{\interfootnotelinepenalty=10000 \footnote{Yearly \emph{Workshop on Asian Translation}, hosting several shared translation tasks. See \url{http://lotus.kuee.kyoto-u.ac.jp/WAT/WAT2019/index.html} and the overview paper \citep{wat19} for the 2019 occurrence.}}, in which different participants, or rather their systems, compete against each other on specific tasks. Subsequently, those test sets are typically also used in research publications to demonstrate the effectiveness of the approach at hand and to compare to previous results. As such, test sets play a crucial role in showing the progress of machine translation. 

For many years, test sets have been prevalently drawn from news articles.\footnote{See  \url{http://matrix.statmt.org/test_sets/list} for example.} 
However, to be able to assess machine translation quality in a wider range of usage scenarios, it is important to also evaluate in other domains than news, and thus to create and establish test sets from a wider range of domains. Clearly, specific usage scenarios have other challenges than what is represented in the news domain. Thus, quality results, and claims about human parity \citep[e.g.][]{hassan-etal-2018}, that have been achieved in the news domain can usually not be directly transferred to other domains. Accordingly, data sets and shared tasks have been created for other domains as well, e.g. biomedical\footnote{See, for example, the biomedical translation task at WMT19: \url{http://www.statmt.org/wmt19/biomedical-translation-task.html}} and patents\footnote{See, for example, the JBO Patent corpus used at WAT: \url{http://lotus.kuee.kyoto-u.ac.jp/WAT/patent/}}. With the \emph{software documentation data set}, we provide the possibility to tune and evaluate \ac{MT} systems in the domain of corporate software documentation, and thus contribute to a clearer picture of the quality of machine translation across domains.
Similarly, the focus of machine translation has often been on high-resource language pairs, such as English-German. With an evaluation data set for four language pairs that are rather on the lower end of availability of resources, we contribute to a better test coverage for the many low-resource language pairs. 

With the recent improvements in machine translation quality, up to claims of human parity, flaws in the evaluation setups and interpretation of results have been pointed out \citep{toral-etal-2018,laubli-etal-2018,wmt18}. Subsequently, more emphasis has been put on carefully evaluating machine translation, in particular to be able to evaluate segments within their document context, e.g.\ by \citet{wmt19}. By creating data sets that consist of documents corresponding to help pages, we contribute to this endeavor. The document structure annotation can also provide additional useful information during human evaluation.
Similarly, machine translation approaches have started to look beyond translating independent sentences. Methods for taking more context into account have emerged, with the goal to improve the translation quality \citep[amongst others]{miculicich-etal-2018,maruf-haffari-2018,lei-2020}. By providing development and test data with document context and metadata, we hope to strengthen such developments. 

Resources that are related to the data set at hand in terms of the covered domain are the data sets from the WMT16 shared task of machine translation of IT domain \citep[][Section 4]{wmt16} and the documentation data set by Salesforce \citep{hashimoto-etal-2019}. The data set from the IT translation shared task consists of answers from a help desk, thus it covers a different text type than software documentation that likely also comes with a different style. Furthermore, the focus of the data set is on European languages, and it does not contain more context than short one-paragraph answers. The data set described and experimented with by \citet{hashimoto-etal-2019} is very similar in nature to ours. Note however that the language scope is different: all language pairs in the data set by Salesforce are rather high-resource.

\section{Origin of the data}
\label{sec:origin}

\subsection{Data sources}
\label{sec:sources}

\begin{figure}[t]
	\centering
	\includegraphics[width=\columnwidth]{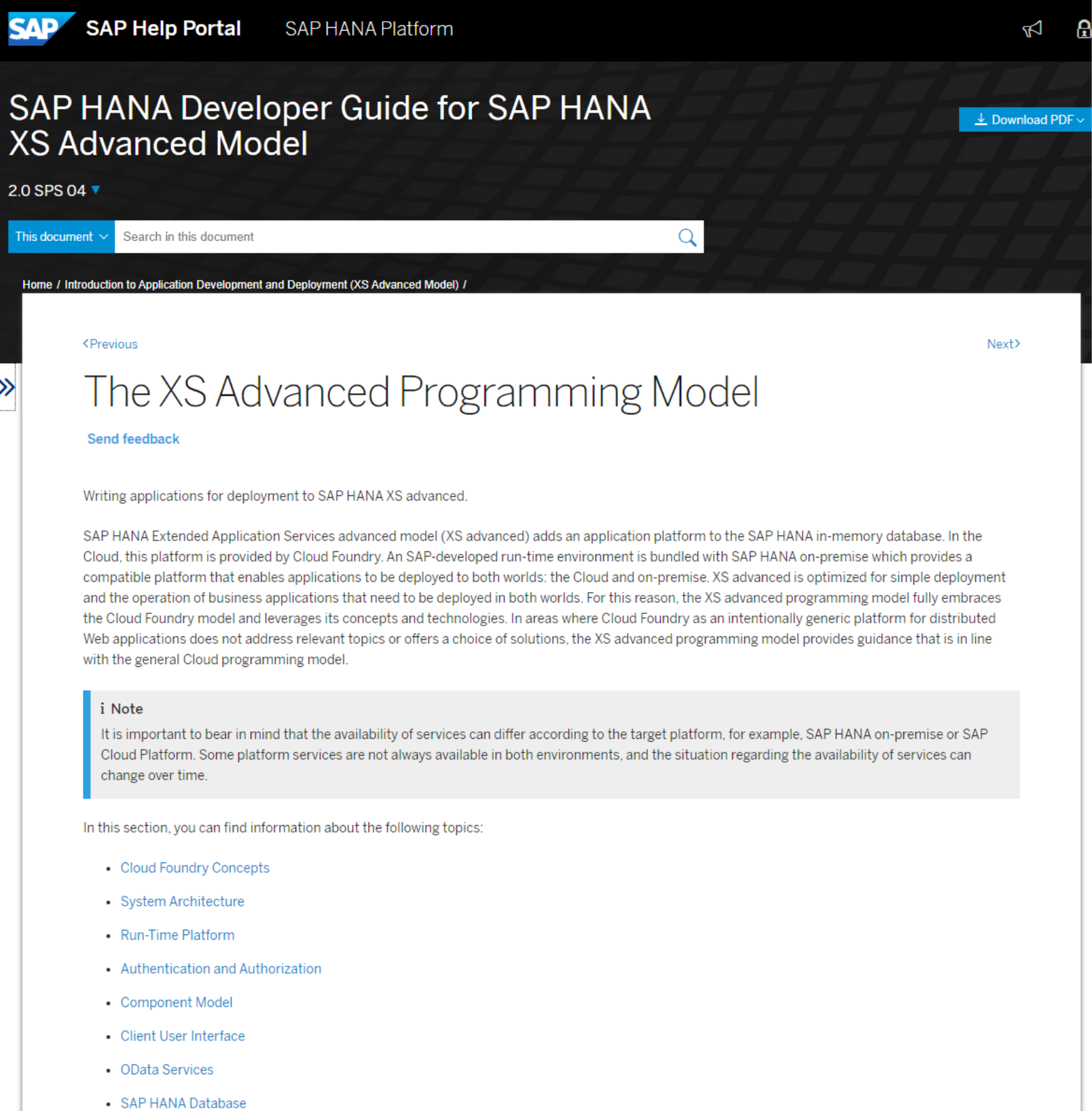}
	\caption{Screenshot of a page from \emph{SAP Help Portal}}
	\label{fig:page-ex}
\end{figure}

The contents of the \emph{software documentation data set for machine translation} originate from the \emph{SAP Help Portal} that contains SAP product documentation and user assistance for product-related questions. 
As it describes the use of software, it is rather technical in nature. In contrast to general textual data, it is highly structured, i.e.\,it contains many tables, lists, links, examples as well as code snippets. 
The textual presentation and page layout follow a similar structure across documents to obtain a coherent appearance of corporate help pages. 
This explains some of the particularities of this data set, described in more detail in Section~\ref{sec:particularities}. 
Figure~\ref{fig:page-ex} shows an example of such a help page. 

The content of the help pages is authored by domain experts and then translated by professional translators that are specialized in the translation of SAP content. Hence, the source data is of excellent quality as well as its translations. Furthermore, the translations of the proposed documentation data set were created without machine translation in the loop, so there is no bias to any \ac{MT} system. 
\subsection{Data preparation}
\label{sec:preparation}

In this section, we will describe the source format of the data and how we processed it for the \emph{software documentation data set for machine translation}. 

English source texts are edited using DITA{\interfootnotelinepenalty=10000 \footnote{\url{https://en.wikipedia.org/wiki/Darwin_Information_Typing_Architecture}}}, an XML-based format, well suited for authoring, structuring and publishing content with a high potential of reuse. For translation, SAP uses \ac{CAT} tools, such as SDL Trados Studio\footnote{\url{https://www.sdl.com/software-and-services/translation-software/sdl-trados-studio/}}. which transform DITA-XML format into XLIFF (XML Localization Interchange File Format)\footnote{\url{http://xml.coverpages.org/xliff.html}} used in translation. As it keeps track of the text structure and inline markup of the source texts, this information can be transferred to the target language after translation. For its use in SDL Trados Studio, SDL developed SDLXLIFF\footnote{\url{http://producthelp.sdl.com/sdl\%20trados\%20studio/client_en/Edit_View/XLIFF_File_Format.htm}}, a special flavor of XLIFF.  SDLXLIFF files are highly structured bilingual files that contain both the source document text and its translation. 

\begin{figure*}[t]
	\centering
	\includegraphics[width=0.95\textwidth]{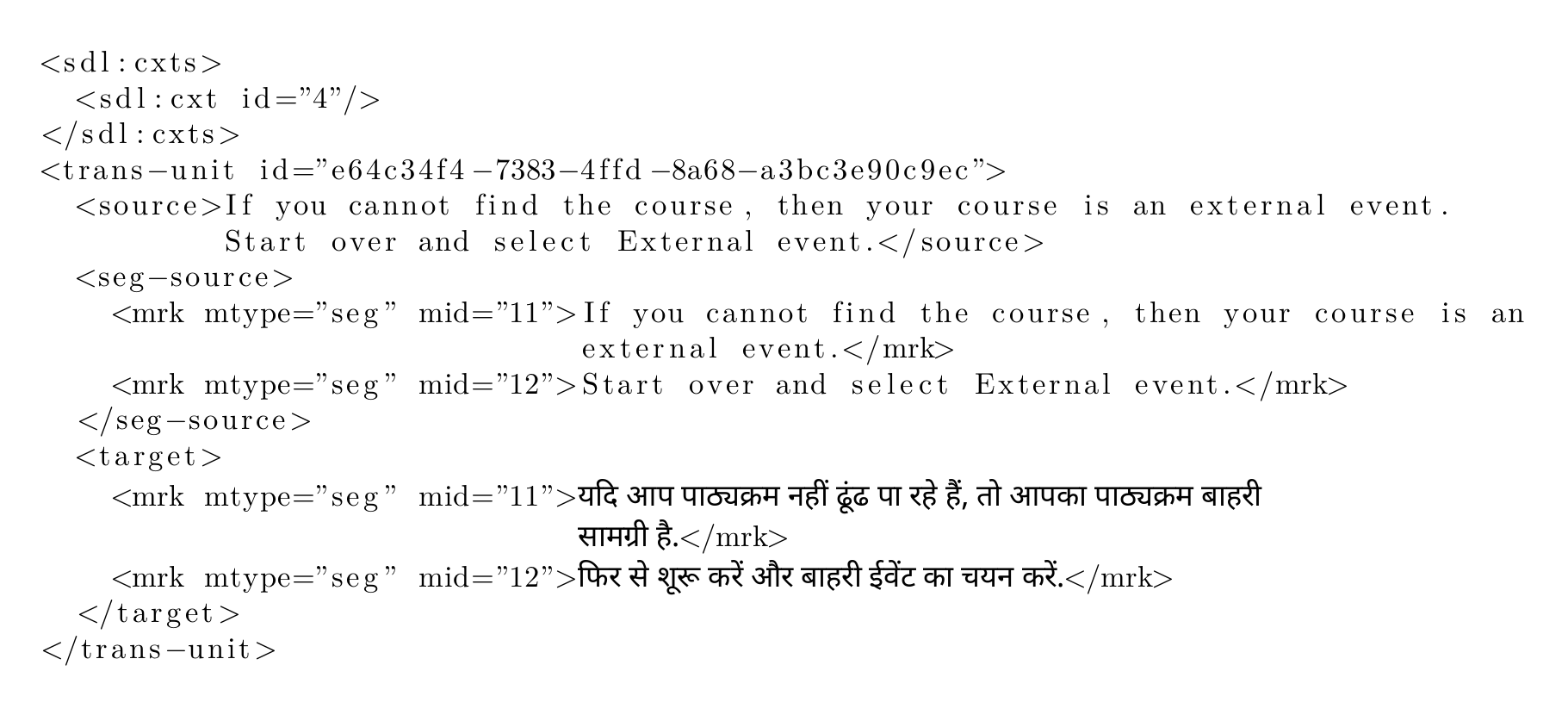}
	\vspace{-5ex}
	\caption{Example of a translation unit in XLIFF format}
	\label{fig:xliff-ex1}
\end{figure*}

Figure~\ref{fig:xliff-ex1} shows a fragment of an SDLXLIFF document that demonstrates the information used to provide parallel text as well as structural annotation of the document context. 
SDLXLIFF files usually cover one document, the content of which is presented in textual order. A translation unit \texttt{<trans-unit>} is a sequence of consecutive text for the source and the target language, in this case for English and Hindi. It is split into sentences by the Trados sentence segmenter, as shown under \texttt{<seg-source>} and \texttt{<target>} in Figure~\ref{fig:xliff-ex1}. Segments are enumerated using the \texttt{mid} attribute. We use this information to order the translation pairs consecutively for each document and to count segments that belong to a text unit or paragraph (see  metadata columns 2 and 4 in Table~\ref{tab:page-ex}).  

\begin{figure*}[t]
	\centering
	\includegraphics[width=0.95\textwidth]{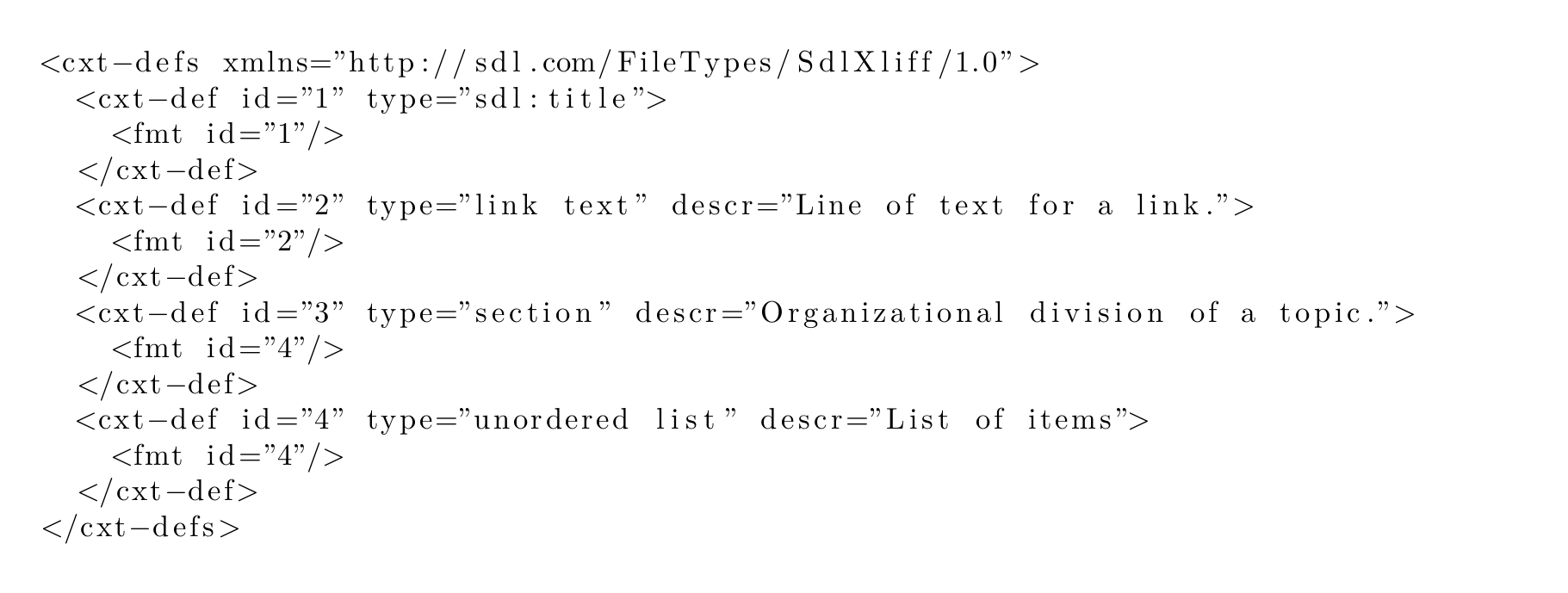}
	\vspace{-5ex}
	\caption{Example of a definition of textual elements in an XLIFF file}
	\label{fig:xliff-ex2}
\end{figure*}

The information about the structural type of a translation unit in the document is conveyed by the \texttt{<sdl:cxts>} context value. Text can be used in a title, a section, a table, an example or an itemized list. In the example in Figure~\ref{fig:xliff-ex1}, the translation unit occurs in the context \texttt{<sdl:cxt id="4"/>} which corresponds to an unordered list, see Figure~\ref{fig:xliff-ex2} for the text element declarations. 

Contextual text types are declared for each XLIFF file and vary depending on the document content and its source. To reduce the number of text types that come with naming variants and different levels of granularity, we mapped them to  six common and self-explanatory categories for the \emph{software documentation data set}: \texttt{title}, \texttt{section}, \texttt{table\_element}, \texttt{list\_element}, \texttt{example}, \texttt{unspecified}. 

Parallel segments, positional metadata and text type were extracted from each SDLXLIFF document using the Saxon parser\footnote{\url{http://saxon.sourceforge.net/}} with an XSLT stylesheet. 
We provide the resulting data in text format, as it is common practice in machine translation, in three sentence-parallel files: source text, target text and document context metadata. The metadata file contains the following five  columns: 
\begin{enumerate}
\item Document ID
\item Segment ID in the document that indicates the contextual order (restarts from 1 in each document)
\item Text Unit ID in the document that indicates segments that occur in consecutive order (starts from 1 in each document). Segments with the same Text Unit ID make up one text block consisting of multiple sentences, for example a paragraph.
\item Segment ID in Text Unit (starts from 1 in each Text Unit)
\item Textual element that describes the structural type of the segment. Values are \texttt{title}, \texttt{section}, \texttt{table\_element}, \texttt{list\_element}, \texttt{example}, \texttt{unspecified}
\end{enumerate}

After the XLIFF processing, the contextual annotation of the content of the SAP Help page in Figure~\ref{fig:page-ex} would look as shown in Table~\ref{tab:page-ex}. It is document 79 with 17 segments and 12 text units. There is a paragraph marked as text unit 3 consisting of 6 sentences. Each list element is considered an individual text unit.\footnote{The \emph{Note} displayed on the help page in Figure~\ref{fig:page-ex} is not part of the document the data was extracted from. It is inserted at some later stage of the publishing process.}

\begin{table*}[t]
\renewcommand{\arraystretch}{1.3}
\centering
\small
\begin{tabular}{p{.6\textwidth}lllll}  
\toprule
\multicolumn{1}{c}{English source} & \multicolumn{5}{c}{Metadata} \\
\cmidrule(r){2-6}
& 1 & 2 & 3 & 4& \multicolumn{1}{c}{5}\\
\midrule
The XS Advanced Programming Model & 79 & 1 & 1 & 1 & \texttt{title} \\
Writing applications for deployment to SAP HANA XS advanced. & 79 & 2 & 2 & 1 & \texttt{section} \\
SAP HANA Extended Application Services advanced model (XS advanced) adds an application platform to the SAP HANA in-memory database. & 79 & 3 & 3 & 1 & \texttt{section} \\
In the Cloud, this platform is provided by Cloud Foundry. &	79 & 4& 3& 2& \texttt{section}\\
An SAP-developed run-time environment is bundled with SAP HANA on-premise which provides a compatible platform that enables applications to be deployed to both worlds: the Cloud and on-premise. &	79& 5& 3& 3& \texttt{section} \\
XS advanced is optimized for simple deployment and the operation of business applications that need to be deployed in both worlds. & 79& 6& 3& 4& \texttt{section} \\
For this reason, the XS advanced programming model fully embraces the Cloud Foundry model and leverages its concepts and technologies.	&79& 7& 3 &5& \texttt{section} \\
In areas where Cloud Foundry as an intentionally generic platform for distributed Web applications does not address relevant topics or offers choice, the XS advanced programming model provides guidance that is in line with the general Cloud programming model.	&79& 8& 3& 6& \texttt{section} \\
In this section, you can find information about the following topics:	&79& 9& 4& 1& \texttt{section} \\
Cloud Foundry Concepts	&79& 10& 5& 1& \texttt{list\_element}\\
System Architecture	&79 &11& 6& 1& \texttt{list\_element}\\
Run-Time Platform	&79& 12& 7& 1& \texttt{list\_element}\\
Authentication and Authorization	&79 &13& 8& 1& \texttt{list\_element}\\
Component Model	 &79& 14& 9& 1& \texttt{list\_element}\\
Client User Interface	&79& 15& 10& 1& \texttt{list\_element}\\
OData Services	&79 &16& 11& 1& \texttt{list\_element}\\
SAP HANA Database	&79& 17& 12& 1 &\texttt{list\_element} \\
\bottomrule
\end{tabular}
\caption{Presentation of source segments and text structure annotation}
\label{tab:page-ex}
\end{table*}
\subsection{Particularities of the data}
\label{sec:particularities}

As pointed out in Section~\ref{sec:sources}, help pages are composed in a way that allows for high reuse of textual content and patterns. For coherent appearance, their structure is intended to be clear and uniform. This has some impact on the kind of text segments we find in software documentation documents. 

\begin{enumerate}
\item There is a lot of \emph{redundancy}, i.e.\,source-target pairs occur several times across documents or even within the same document. This concerns titles, table headers, table values or even complete sentences. 
\item As the help pages on the \emph{SAP Help Portal} contain many tables and list items, many \emph{translation segments are short}, sometimes consisting of just a number or one word. 
 List and table elements are presented as individual text units and are translated independently within their document context. 
\item There is a large number of \emph{short documents} reflecting the segmentation of help page content into reusable units. 
\end{enumerate}

These particularities impact the creation of the evaluation data sets (Section~\ref{sec:selection}) and the characteristics of the final data set (Section~\ref{sec:characteristics}).

\subsection{Data selection}
\label{sec:selection}

Ideally, test and development sets for machine translation meet the requirements of being\footnote{General guidance for assembling (test) data can be found in \citet[Sec.~1.6.5]{megerdoomian}, \citet[Sec.~4.3]{jm}, \citet[Sec.~2.6]{eval-nlp}, amongst others.} (i)~ \emph{representative} for a given test or usage scenario, in our case for a given domain, covering well its specific terminology, its syntax and style,
(ii)~\emph{free of duplicates} and redundancy,
(iii)~\emph{balanced}, i.e., ideally sampled from a larger set of data, so that the content is spread over various topics.

When building evaluation sets as collections of single sentences (or sentence pairs), it is rather straightforward to adhere to these criteria. However, when creating them for whole documents, the absence of duplicates and redundancy as well as content balance are more challenging. This is particularly true for our help page content that displays similar structuring and repetitions, see Section~\ref{sec:particularities}. Obviously, duplicate sentence pairs cannot simply be removed if we want to keep the contextual order of segments. 

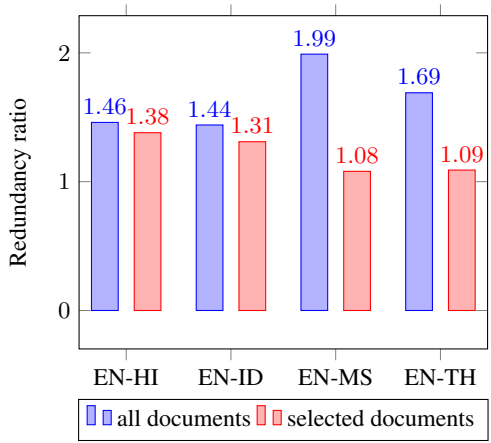
\begin{figure}[t]
\centering
\small
\begin{tikzpicture}
\begin{axis}[
    ybar=6pt,
    height=6cm,
    ymin=0,
    enlargelimits=0.15,
    legend style={at={(0.5,-0.15)},
      anchor=north,legend columns=-1},
    ylabel={Redundancy ratio},
    y label style={yshift=-13pt},
    symbolic x coords={\en-\hi,\en-\id,\en-\ms,\en-\thai},
    xtick=data,
    nodes near coords,
    nodes near coords align={vertical},
    ]
\addplot coordinates {(\en-\hi,1.46) (\en-\id,1.44) (\en-\ms,1.99) (\en-\thai,1.69)};
\addplot coordinates {(\en-\hi,1.38) (\en-\id,1.31) (\en-\ms,1.08) (\en-\thai,1.09)};
\legend{all documents,selected documents}
\end{axis}
\end{tikzpicture}
	\caption{Redundancy reduction: redundancy in all data vs. the data that was selected for the data set.}
	\label{fig:redundancy}
\end{figure}

Let us define \emph{redundancy} as the ratio of all source-target pairs to unique source-target pairs in a data set. 
Figure~\ref{fig:redundancy} shows the redundancy for all data at our disposal (in blue).
We see that it differs depending on the language pair. 
To some extent, this can be explained by the amount of documents used for extraction. While for English  to  Malay (\en-\ms) and to Thai (\en-\thai) we had several thousands of original documents at hand, for English to Hindi (\en-\hi) and to Indonesian (\en-\id) only a couple of hundred documents were available that had less overlap and thus show less redundancy.


To meet the requirements of test and development data, we made an effort to reduce this redundancy by selecting documents that are less prone to have content present in other documents. 
The following indicators were calculated to be used in the selection process:
\begin{itemize}
\item Document redundancy ratio: percentage of unique parallel segments to all parallel segments in a document (to flag documents that contain duplicates).
\item Number of segments in the document (to flag documents with little content, and hence context).
\item Average number of source words per segment (to keep documents with longer segments).
\item Cross-document redundancy of a document with respect to all documents (to flag documents that contain a large number of segments that occur in many other documents). We first created a frequency list of source segments of all documents. Then, for each segment of a document, their overall document frequencies were summed up and divided by the number of segments in the document. This ratio is high if the document contains many segments that occur in many documents.  
\item Document double indicator. It turned out that for \en-\ms\ and \en-\thai, many documents were almost identical but for one or two segments. Overall cross-document redundancy does not help in this case, as source-target pairs occur only twice. The document double indicator flags documents that contain a large percentage of source-target pairs that occur exactly twice in the complete data.
\end{itemize}

For each language pair, we selected a subset of all available documents that contains about 4k sentences that meets the requirements as much as possible by calibrating the indicators. 
For \en-\ms\ and \en-\thai, all five indicators were used to reduce the redundancy as much as possible.  For \en-\hi\ and \en-\id, only the document redundancy ratio and the number of segments per document were considered, as there were less documents to choose from and there was less redundancy to start with. With this approach, we successfully obtained a data set with less duplicates across documents, see Figure~\ref{fig:redundancy} (in red).

\section{Characteristics of the data set}
\label{sec:characteristics}

The selected documents with reduced redundancy, see Section~\ref{sec:selection}, were divided into development and test data sets and constitute the \emph{software documentation data set for machine translation}. We will look into its characteristics in this section. Table~\ref{tab:stats} provides an overview over the size of and redundancy within the respective data set. 


\begin{table*}[p!]
\centering
\small
\begin{tabular}{lcccccccc}  
\toprule
& \multicolumn{2}{c}{\# of documents} & \multicolumn{2}{c}{\# of parallel segments} & \multicolumn{2}{c}{\# of source words} & \multicolumn{2}{c}{Data set redundancy}\\
\cmidrule(lr){2-3} \cmidrule(lr){4-5} \cmidrule(lr){6-7} \cmidrule(lr){8-9}
& dev & test & dev & test & dev & test & dev & test \\
\midrule
\en-\hi & 78	&76	&2,016	&2,073	&20,662	&18,128 & 1.33 &1.14\\
\en-\id	&66	&74	&2,023	&2,037	&21,159	&18,164 & 1.26 & 1.11\\
\en-\ms &210	&197	&2,050	&2,050	&26,654	&26,758 & 1.04 & 1.05\\
\en-\thai	&207&	205	&2,048	&2,050	&25,759&	25,426 & 1.03 & 1.05\\
\bottomrule
\end{tabular}
\caption{Statistics on development and test data sets}
\label{tab:stats}
\end{table*}

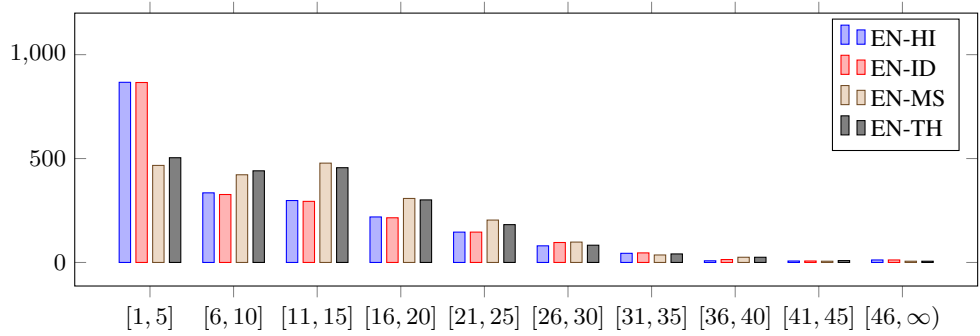
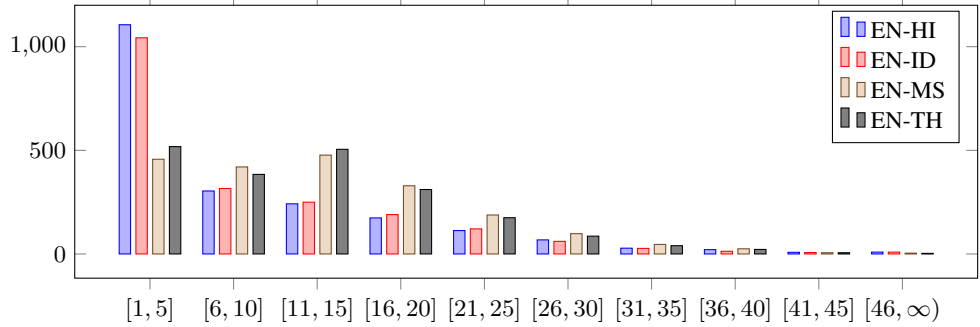
\begin{figure*}[p!]
\centering
\small

\begin{subfigure}{\textwidth}
\centering
\begin{tikzpicture}
\begin{axis}[
    ybar,
    bar width=0.15cm,
    x=1.1cm,
    ymax=1200,
    height=5.2cm,
    legend cell align={left},
    symbolic x coords={1,2,3,4,5,6,7,8,9,10},
    xticklabels={$\interval{1}{5}$,$\interval{6}{10}$,$\interval{11}{15}$,$\interval{16}{20}$,$\interval{21}{25}$,$\interval{26}{30}$,$\interval{31}{35}$,$\interval{36}{40}$,$\interval{41}{45}$,$\interval[open right]{46}{\infty}$},
    xtick=data,
    ]
\addplot coordinates {(1,867) (2,335) (3,298) (4,219) (5,146) (6,80) (7,44) (8,8) (9,7) (10,12)};
\addplot coordinates {(1,866) (2,327) (3,294) (4,215) (5,146) (6,96) (7,46) (8,14) (9,7) (10,12)};
\addplot coordinates {(1,467) (2,422) (3,478) (4,308) (5,204) (6,98) (7,36) (8,25) (9,6) (10,6)};
\addplot coordinates {(1,504) (2,441) (3,456) (4,301) (5,182) (6,83) (7,41) (8,25) (9,9) (10,6)};
\legend{\en-\hi,\en-\id,\en-\ms,\en-\thai}
\end{axis}
\end{tikzpicture}
\caption{Development data}
\end{subfigure}

\begin{subfigure}{\textwidth}
\centering
\begin{tikzpicture}
\begin{axis}[
    ybar,
    bar width=0.15cm,
    x=1.1cm,
    ymax=1200,
    height=5.2cm,
    legend cell align={left},
    symbolic x coords={1,2,3,4,5,6,7,8,9,10},
    xticklabels={$\interval{1}{5}$,$\interval{6}{10}$,$\interval{11}{15}$,$\interval{16}{20}$,$\interval{21}{25}$,$\interval{26}{30}$,$\interval{31}{35}$,$\interval{36}{40}$,$\interval{41}{45}$,$\interval[open right]{46}{\infty}$},
    xtick=data,
    ]
\addplot coordinates {(1,1106) (2,304) (3,242) (4,174) (5,113) (6,68) (7,28) (8,21) (9,8) (10,9)};
\addplot coordinates {(1,1043) (2,316) (3,250) (4,190) (5,121) (6,61) (7,27) (8,13) (9,7) (10,9)};
\addplot coordinates {(1,457) (2,420) (3,477) (4,329) (5,188) (6,98) (7,46) (8,25) (9,6) (10,4)};
\addplot coordinates {(1,518) (2,384) (3,505) (4,311) (5,175) (6,86) (7,40) (8,22) (9,6) (10,3)};
\legend{\en-\hi,\en-\id,\en-\ms,\en-\thai}
\end{axis}
\end{tikzpicture}
\caption{Test data}
\end{subfigure}

	\caption{Length distributions of source segments}
	\label{fig:length}
\end{figure*}

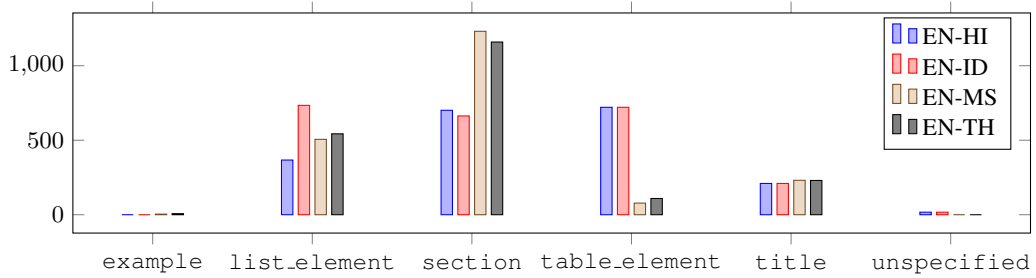
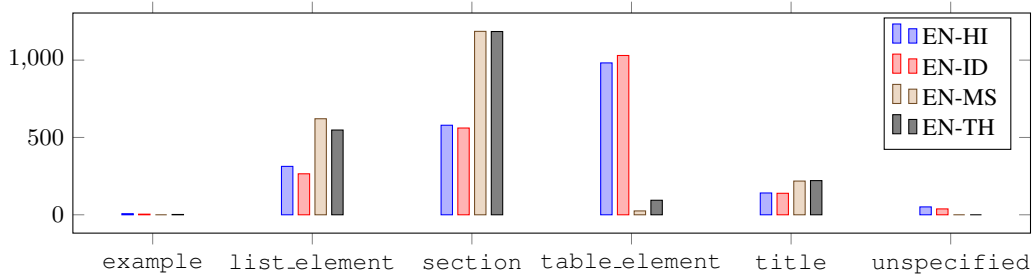
\begin{figure*}[p!]
\centering
\small

\begin{subfigure}{\textwidth}
\centering
\begin{tikzpicture}
\begin{axis}[
    ybar,
    bar width=0.15cm,
    x=2.1cm,
    height=4.5cm,
    legend cell align={left},
    symbolic x coords={1,2,3,4,5,6},
    xticklabels={\texttt{example}, \texttt{list\_element}, \texttt{section}, \texttt{table\_element}, \texttt{title}, \texttt{unspecified}},
    xtick=data,
    ]
\addplot coordinates {(1,0) (2,367) (3,701) (4,721) (5,210) (6,17)};
\addplot coordinates {(1,0) (2,734) (3,663) (4,721) (5,210) (6,17)};
\addplot coordinates {(1,4) (2,506) (3,1231) (4,78) (5,231) (6,0)};
\addplot coordinates {(1,7) (2,543) (3,1159) (4,109) (5,230) (6,0)};
\legend{\en-\hi,\en-\id,\en-\ms,\en-\thai}
\end{axis}
\end{tikzpicture}
\caption{Development data}
\end{subfigure}

\begin{subfigure}{\textwidth}
\centering
\begin{tikzpicture}
\begin{axis}[
    ybar,
    bar width=0.15cm,
    x=2.1cm,
    height=4.5cm,
    legend cell align={left},
    symbolic x coords={1,2,3,4,5,6},
    xticklabels={\texttt{example}, \texttt{list\_element}, \texttt{section}, \texttt{table\_element}, \texttt{title}, \texttt{unspecified}},
    xtick=data,
    ]
\addplot coordinates {(1,7) (2,313) (3,579) (4,982) (5,141) (6,51)};
\addplot coordinates {(1,4) (2,265) (3,561) (4,1030) (5,139) (6,38)};
\addplot coordinates {(1,0) (2,621) (3,1186) (4,25) (5,218) (6,0)};
\addplot coordinates {(1,2) (2,548) (3,1185) (4,94) (5,221) (6,0)};
\legend{\en-\hi,\en-\id,\en-\ms,\en-\thai}
\end{axis}
\end{tikzpicture}
\caption{Test data}
\end{subfigure}

	\caption{Distribution of textual element annotations}
	\label{fig:anno}
\end{figure*}

While the number of segments of the development and test sets are in the same range across language pairs, the number of documents and the total amount of words are different for \en-\hi\ and \en-\id\  compared to the other two language pairs. This difference is also reflected in the distribution of words per segment, see Figure~\ref{fig:length}: there is a larger number of short segments for \en-\hi\ and \en-\id. For  \en-\ms\ and \en-\thai, we see a more balanced distribution of short and medium length segments in both, development and test sets.
Figure~\ref{fig:anno} shows the distributions of textual element annotations in the data sets’ metadata. They explain, to some extent, the distribution of segment length:  We see a larger number of \texttt{section} segments for \en-\ms\ and \en-\thai. Sections usually contain longer segments than table elements, which are frequent for \en-\hi\ and \en-\id.

Finally, we look at the redundancy in the released data set, i.e.\ the number of all source-target pairs over the number of unique source-target pairs, shown in Table~\ref{tab:stats}. As expected from Figure~\ref{fig:redundancy}, there is more redundancy for \en-\hi\ and \en-\id, which ties in with the larger number of shorter segments. They are more likely to reoccur across documents.  

In summary, we conclude that the data sets for \en-\hi\ and \en-\id\ are comparable concerning the criteria analyzed in this section. They differ somewhat from the \en-\ms\ and \en-\thai\ data sets that also have characteristics in common. We would have preferred to provide a more homogeneous data set. However, given the different sizes and features of the original resources and the constraints imposed by adding contextual metadata, this was not feasible.
On the other hand, the charts in this section indicate that the development and test sets of each language pair share the same characteristics, i.e.\ their segment length distribution, the types of textual elements as well as their word counts are comparable. This makes the development sets well suited to optimize an \ac{MT} model towards the translation of the corresponding test set.

\section{Machine translation results}
\label{sec:mt}

In this section, we provide reference \ac{MT} results on the test sets of the \emph{software documentation data set for machine translation}. They should serve as comparison for future research evaluated on this data set. The results can be found in Table~\ref{tab:mt}.


\begin{table}
\centering
\small
\begin{tabular}{lccccc}  
\toprule
& Provider & BLEU & ChrF  & BLEU$^\ast$ \\
\midrule
\multirow{2}{*}{\en-\hi} & WAT & 13.67 & 0.3193  & 13.73 \\
& Online & 32.97 & 0.5681 \\
\multirow{2}{*}{\en-\id} & WAT & 30.39 & 0.5828 & 30.11 \\
& Online & 58.18 & 0.7662 \\
\multirow{2}{*}{\en-\ms} & WAT & 31.85 & 0.5968 & 31.95 \\
& Online & 42.31 & 0.7005 \\
\multirow{2}{*}{\en-\thai} & WAT & 31.29 & 0.2933 & 31.28  \\
& Online & 68.80 & 0.6443 \\
\midrule
\multirow{2}{*}{\hi-\en} & WAT & 14.54 & 0.3987 & 14.39 \\
& Online & 50.19 & 0.7375 \\
\multirow{2}{*}{\id-\en} & WAT & 23.25 & 0.4917 & 23.05 \\
& Online & 52.94 & 0.7552 \\
\multirow{2}{*}{\ms-\en} & WAT & 25.32 & 0.5120 & 25.36 \\
& Online & 48.52 & 0.7313 \\
\multirow{2}{*}{\thai-\en} & WAT & 9.56 & 0.3244 & 9.56 \\
& Online & 27.73 & 0.5717 \\
\bottomrule
\end{tabular}
\caption{Machine translation results on the test set}
\label{tab:mt}
\end{table}

The first reference \ac{MT} result is based on the baseline system of the WAT 2020 \emph{NICT-SAP IT and Wikinews Task}\footnote{See \url{http://lotus.kuee.kyoto-u.ac.jp/WAT/NICT-SAP-Task/} for the details.}, provided by the organizers of the task. There are two multilingual systems, one for \en\ to \{\hi, \id, \ms, \thai\} and one for \{\hi, \id, \ms, \thai\} to \en. They are trained on Wikinews data from the Asian Language Treebank (ALT) project \citep{alt} and IT data from Opus\footnote{\url{http://opus.nlpl.eu/}} (Ubuntu, GNOME and KDE4) \citep{tiedemann-2012}. A transformer-big configuration \citep{vaswani-etal-2017} was used. More details can be found in the workshop overview \citep{wat20}. The second reference \ac{MT} system that we report on is a popular general purpose online \ac{MT} provider. The contextual metadata of the test set is not used in either reference system.

We report case-sensitive BLEU \citep{bleu} and ChrF \citep{chrf2} scores as calculcated by \emph{sacrebleu}{\interfootnotelinepenalty=10000 \footnote{Version strings are \texttt{BLEU+case.mixed+numrefs.1+} \texttt{smooth.exp+tok.13a+version.1.4.13} and  \texttt{chrF2+numchars.6+space.false+} \texttt{version.1.4.13}.}} \citep{sacrebleu}. The BLEU scores for \en-\thai\  are generated based on character-segmented input. For the WAT system, we additionally report the official BLEU scores as provided by the task evaluation (cf. column BLEU$^\ast$ in Table~\ref{tab:mt}).\footnote{See the \texttt{SOFTWARE*} links on \url{http://lotus.kuee.kyoto-u.ac.jp/WAT/evaluation/index.html}. Pages accessed on September 15th, 2020.} It uses \emph{Moses' multi-bleu.perl}.\footnote{See \url{http://lotus.kuee.kyoto-u.ac.jp/WAT/evaluation/automatic_evaluation_systems/tools.html} for details on the used tools.}

The online \ac{MT} provider outperforms the WAT baseline for all language pairs by a wide margin on our test sets. Given the low-resource setting of the baseline WAT system and its role as a simple baseline, this is not very surprising. 
We expect that by using more data, in particular in-domain data, may it be monolingual or parallel, the gap could be narrowed or even closed.
Other interesting questions are whether bilingual systems would perform better than the multilingual baseline, how to better exploit small quantities of in-domain data, how to leverage the available contextual metadata, and whether neural network parametrization could help to improve results in this low-resource setting.
\section{Conclusion}
\label{sec:concl}


We released the \emph{software documentation data set for machine translation} to the MT research community, a high-quality real-world data set with content from the \emph{SAP Help Portal}. To our knowledge, it is the first data collection with explicit text structure annotation and the first IT-specific evaluation data set for English to Hindi, Indonesian, Malay and Thai. It will advance automatic quality assessment of context-aware MT systems, giving users the flexibility to consider all or only selected or no text structure metadata. Moreover, it facilitates the development and testing of machine translation systems for low-resource language pairs for the translation of software documentation in a corporate context.
As a starting point, we provided MT results that serve as benchmarks in future research.

\pagebreak
\section*{Acknowledgments}
We would like to thank Raj Dabre, Masao Utiyama and Eiichiro Sumita from NICT for their support, for training the baseline WAT systems and for providing us with the translations. We would also like to thank our colleagues Dominic Jehle and Matthias Huck for comments on the draft.

\bibliography{references}
\bibliographystyle{acl_natbib}

\end{document}